\documentclass[11pt,a4paper]{article}
\usepackage[hyperref]{acl2021}
\usepackage{times}
\usepackage{latexsym}

\usepackage{microtype}

\aclfinalcopy 

\usepackage{ amssymb }

\usepackage{array}
\newcolumntype{\myvrule}[1]{!{\vrule width #1}}

 \usepackage{tabularx}

\usepackage{linguex}

\usepackage{amsmath}

\usepackage[T1]{fontenc}

\usepackage{kevinmacros}
\renewcommand{\mycitet}[1]{~\citet{#1}}           
\renewcommand{\mycitep}[1]{~\citep{#1}}       		
\renewcommand{\meto}[1]{\textbf{{#1}}}           

\newcommand{\negpercent}{15}

\usepackage{graphicx, import}
\graphicspath{ {images/} }

\usepackage{subcaption}

\usepackage{blindtext}

\usepackage{paralist}

\usepackage{multirow}

\usepackage{tikz}
\usetikzlibrary{arrows.meta}
\tikzset{%
  	>={Latex[width=2mm,length=2mm]},
    tokenBlock/.style = {
    		rectangle, rounded corners, draw=black,
              minimum width=1.8cm, minimum height=1cm,
              text centered, fill=green!30},
    smallBlock/.style = {
    		rectangle, rounded corners, draw=black,
              minimum width=1.8cm, minimum height=1cm,
              text centered, fill=yellow!30},
    longBlock/.style = {
    		rectangle, rounded corners, draw=black,
              minimum width=1.8\linewidth, minimum height=1cm,
              text centered, },
    superlongBlock/.style = {
    		rectangle, rounded corners, draw=black,
              minimum width=2.09\linewidth, minimum height=1cm,
              text centered,},
    midBlock/.style = {
    		rectangle, rounded corners, draw=black,
              minimum width=1.02\linewidth, minimum height=1cm,
              text centered, },
    mid2Block/.style = {
    		rectangle, rounded corners, draw=black,
              minimum width=0.76\linewidth, minimum height=1cm,
              text centered,},
    posLabelBlock/.style = {
    		rectangle, rounded corners, draw=black,
              minimum width=1.8cm, minimum height=1cm,
              text centered, fill=red!30},
    negLabelBlock/.style = {
    		rectangle, rounded corners,
              minimum width=1.8cm, minimum height=1cm,
              text centered,},
    stealthBlock/.style = {
    		rectangle, rounded corners,
              minimum width=1.8cm, minimum height=1cm,
              text centered,},
	every node/.style = {
			fill=white, font=\Large}
}
\usepackage{tikz-qtree}
\usetikzlibrary{arrows,positioning}

\title{Impact of Target Word and Context on End-to-End Metonymy Detection}

\author{Kevin Alex Mathews 
\qquad
  Michael Strube \\
  Heidelberg Institute for Theoretical Studies \\
  Schloss-Wolfsbrunnenweg 35, 69118 Heidelberg, Germany \\
  \texttt{\{kevin.mathews, michael.strube\}@h-its.org} \\}

\date{}

\begin{document}

\maketitle
\begin{abstract}
Metonymy 
	is a figure of speech
	in which an entity is referred to 
	by another related entity.
The task of metonymy detection
	aims to distinguish 
	metonymic tokens from literal ones.
Until now,
	metonymy detection methods
	attempt to disambiguate
	only a single noun phrase in a sentence, 
	typically 
	location names 
	or 
	organization names.
In this paper, 
	we disambiguate
	every word in a sentence
	by reformulating
	metonymy detection
	as a sequence labeling task.
We
	also
	investigate the impact of target word and context
	on metonymy detection.
We show that
	the target word
	is less useful
	for detecting metonymy
	in our dataset.
On the other hand,
	the entity types
	that are associated with
	domain-specific words
	in their context
	are easier to solve.
This shows that
	the context words are
	much more relevant
	for detecting metonymy.
\end{abstract}

\section{Introduction}\label{section:introduction}
	Metonymy is a figure of speech
		in which an entity is referred to by another related entity
		\mycitep{lakoff1980metaphors,littlemore_2015}.
	Note how 
		the word \meto{Barcelona} 
		is interpreted in the following two snippets of text:

	\ex.	Toral was born in \meto{Barcelona}, in the Province of Tarragona, Catalonia. He began playing football as a child with his local club, UE Barri Santes Creus.
	 \label{example:ex-lit}

	\ex.  Modri{\'c} started in Real Madrid's home match against rivals \meto{Barcelona}. From a corner kick, he assisted Sergio Ramos to score the winning goal, giving Real a victory in El Cl{\'a}sico.
	\label{example:ex-met}

	In the former, 
		\meto{Barcelona} refers to the city of Barcelona located in the country of Spain.
	However, 
		in the latter,
		the same word does not refer to the city of Barcelona,
		but to a sports team 
		as is evident from 
		the context words
		such as match, rivals,  goal and victory.
	The 	
		likely \emph{hidden} entity here
		is the football club FC Barcelona based in the city of Barcelona.
%
	The linguistic phenomenon 
		in action here is metonymy.
	Although the word \meto{Barcelona} refers to the city in its general (or literal) sense, 
		the same word refers to the football team, a different but related entity, in its metonymic sense.

	Metonymy is frequent in 
		verbal as well as 
		written communication.
	According to\mycitet{P17-1115},
		the natural distribution
		of literal and metonymic usage
		is 80\% and 20\% respectively 
		based on location names
		sampled from \wikipedia.
	Resolving metonymy in text
		aids 
		natural language processing (\nlp) tasks
		such as
 		entity linking\mycitep{ling-etal-2015-design}, 
 		coreference resolution\mycitep{recasens-etal-2010-typology},
		geoparsing\mycitep{gritta-etal-2018-melbourne},
 		and 
		may also aid
 		machine translation\mycitep{MARKERT2002145}.
	In spite of this, 
		metonymy,
		as opposed to 
		metaphor\mycitep{mao-etal-2019-end}
		or
		sarcasm\mycitep{khodak-etal-2018-large},		
		has not received
		much attention
		in the \nlp community.

	The task of metonymy detection
		aims to
		identify metonymy in text.
%
	In this paper, 
		we 
		present
		and 
		compare
		different models for
		metonymy detection.
	In the task as performed in previous work{\mycitep{li2020target,P17-1115,Nastase:2012:LGC:2390948.2390971},
		a specific noun phrase
		in a sentence
		is designated for disambiguation.
	The target,
		known as
		the potentially metonymic word (\pmw),
		is
		typically
		a mention such as 
		a location name or an organization name.
	However, 
		such target words
		are not highlighted
		in real-world texts.
	In addition, 
		if the \pmw is labeled in advance, 
		the models perform very well.
	For instance,
		the \bert-based metonymy detection system,
		proposed by \mycitet{li2020target},
		reports an accuracy of 95.9
		on the \wimcor\mycitep{lrec20-wimcor} data.

	In this paper,
		we 
		consider every word
		to be 
		potentially metonymic
		and 
		disambiguate
		every word in a sentence.
	As a result, 
		metonymy detection
		can readily be applied
		to different
		downstream tasks
		without any fine-tuning.
	For this purpose, 
		we reformulate
		metonymy detection
		as a sequence labeling task.
	This setting
		realizes
		end-to-end metonymy detection.
	Formally
		we define the task of metonymy detection
		as follows:
		given a sequence of tokens ($t_{1}$, $t_{2}$, $t_{3}$, $\dots$, $t_{n}$),
		predict a sequence of labels ($\ell_{1}$, $\ell_{2}$, $\ell_{3}$, $\dots$, $\ell_{n}$)
		to indicate 
		whether a word is
		metonymic or not.

	In addition,\mycitet{li2020target} 
		observes
		that
		the interpretation of a target word
		relies more on the context than the word itself.
	To test 
		this claim
		for end-to-end metonymy detection,
		we compare
		two variants:
		one variant
		relies primarily on the target word,
		and
		the other variant
		relies primarily on the context.
	Our results
		show that masking the target word
		improves end-to-end metonymy detection, 
		especially in the fine-grained experimental setting.

	Note that\mycitet{li2020target} 	
		use their proposed \bert-based model
		for end-to-end metonymy resolution.
	However
		they use \bert named entity recogniser
		to detect locations
		and then these location names are checked for metonymy.
	Our method does not use 
		any such external resource.

	In short, our contributions 
		are as follows:
	\begin{inparaenum}[(1)]
		\item formulate metonymy detection as a sequence labeling task (end-to-end metonymy detection),
		\item adapt an existing dataset for sequence labeling, and
		\item compare the impact of target word and context words.
	\end{inparaenum}

\section{Related Work}
	\subsection{Metonymy Resolution}
	
 of classes.
				
	Several studies
		have explored
		different hand-crafted features 
		for metonymy detection
		such as
		co-occurrence, collocation and grammatical features\mycitep{MARKERT2002145,Markert:2002:MRC:1118693.1118720,Nissim:2003:SFW:1075096.1075104},
		associative information\mycitep{TERAOKA2011105,Teraoka2016MetonymyAU},
		and
		selectional preference features\mycitep{Nastase:2009:CCL:1699571.1699631,Nastase:2012:LGC:2390948.2390971}.

	\mycitet{P17-1115}
		employ a neural-network-based model
		that
		uses the context words 
		of the \pmw
		to identify whether
		the \pmw is literal or metonymic.	
	\mycitet{lrec20-wimcor}
		propose
		\wiki,
		a larger and richer dataset
		of location metonymy,
		extracted 
		using \wikipedia.
	They
		construct benchmarks
		for metonymy detection
		using \glove and \bert embeddings.
	\mycitet{li2020target} 
		propose target word masking 
		to use \bert for metonymy detection.	
	For this purpose, 
		they mask target words
		during training and testing.
	This model
		outperforms
		the model
		that sees the target word
		during detection.


	\subsection{Word Sense Disambiguation}
	
	Both metonymy resolution
		and
		word sense disambiguation\mycitep{10.1145/1459352.1459355}
		deal with lexical ambiguity.
	The principal difference
		is that
		the candidate interpretations
		of a potentially metonymic word
		are strongly related
		to each other.
	For instance, 
		Barcelona
		and 
		FC Barcelona
		form a metonymic pair of candidates
		because 
		the football club is based in the city
		and
		both entities can be referred to by the same word \emph{Barcelona}.

	On the other hand,
		word sense disambiguation
		primarily deals with
		other linguistic phenomena 
		such as
		 polysemy or homonymy.
	While 
		polysemy pertains to a textual item
		having multiple fine-grained \emph{related} senses,
		homonymy pertains to two (or more) textual items 
		that are \emph{accidentally} 
		similar in surface form. 
	The city of Paris in France
		and
		Paris, the mythological character,
		fail to form a metonymic pair of candidates
		because of the lack of any 
		strong relationship between the two entities,
		although both can be referred to by the same word \emph{Paris} through their association via homonymy.

	According to\mycitet{alan2000chapter},		
		metonymy, along with metaphor, is a type of non-linear polysemy
		due to the non-linear (non-taxonomic) relationship 
		between the literal (most-common) interpretation and the figurative interpretation.

	According to\mycitet{adam2006chapter}, 
		it is difficult to generate a inventory of word senses
		due to the various \emph{policies} to make such as
	 	which senses to be merged, 
		which senses to be considered distinct
		or
		how to address issues such as metaphor, metonymy.
	Lexicographers rely on corpus to identify different word senses
		and 
		thus typically new word senses originate slowly.

	According to the ICM view of metonymy, 
		anything that is related can be a potential candidate; even new entities.
	So the candidate senses of metonymy resolution
		are more open-ended\mycitep{nunberg1978}.

	Metonymy resolution
		has been influenced
		by advances in word sense disambiguation
		in the selection
		of hand-crafted features\mycitep{Markert:2002:MRC:1118693.1118720}
		and 
		techniques\mycitep{Nastase:2009:CCL:1699571.1699631}.
	The fine-grained evaluation
		discussed in this paper
		is closely related to
		word sense disambiguation
		where
		the candidate senses are different entity types.
				
%
%

	\section{Methodology}

	Pretrained language models
		based on transformers\mycitep{NIPS2017_3f5ee243}
		are successfully used in various 
		natural language understanding tasks\mycitep{wu2019zero,baldini-soares-etal-2019-matching}.
	We deploy a pretrained language model
		and then fine-tune the model
		for metonymy detection.
	To fine-tune the model,
		we add a recurrent layer,
		and 
		a fully-connected layer
		on top of the final encoder layer.

	We create two variants
		by feeding
		different input to the recurrent layer, 
		as described below:
	\begin{asparaenum}[(1)]
		\item 
		In the first variant,
			the vector representation of the current token is given as input at each timestep.
		We refer to this model
			as \seqlab (\figref{fig:nomaskmodel}).
		\item 
		In the second variant,
			the vector representations of the context words of the current token 
			are given as input 
			to predict the label of the current token.
		We refer to this model
			as \seqlabbasen{n} (\figref{fig:maskmodel}),
			where \emph{n} refers to the length of the context window.
	\end{asparaenum}

	The performance of the former
		is highly tied 
		to the vector representation 
		of the target word.
	The latter
		relies more on the context
		to detect metonymy.
	Having these two variants
		makes it possible 
		to compare the 
		impact of presence of the target word
		and context
		on end-to-end metonymy detection.	

	The model architecture
		of \seqlabbasen{n}
		is inspired by\mycitet{P17-1115}, 
		whose model
		performed well
		on the metonymy datasets
		\semeval\mycitep{markert-nissim-2007-semeval}
		and 
		\relocar\mycitep{P17-1115}.
	Their model
		used
		a predicate window of context words, 
		which is a small and focused context 
		formed using the grammatical head of the target word.
	However, in this work, 
		we use the adjacent words as context
		in order to avoid the overhead necessary 
		to compute the predicate window
		of every token.

		
%
	In our experiments, 
		we use the pretrained base, uncased version of 
		\bert\mycitep{devlin-etal-2019-bert}.
	We add the representations 
		from the last four hidden layers of the 
		\bert transformer 
		to compute subword embeddings.
	According to \mycitet{devlin-etal-2019-bert},
		the only other method
		that outperforms summation
		is concatenation of last four hidden layers.
	However concatenation
		creates longer vectors
		and thus
		increases memory requirements.
	These subword embeddings are then combined
		through summation 
		to generate the word embeddings.
	In this way,
		we ensure that 
		the feature length of a sequence 
		matches 
		its label length.
	The gradients
		are not backpropagated 
		into \bert.

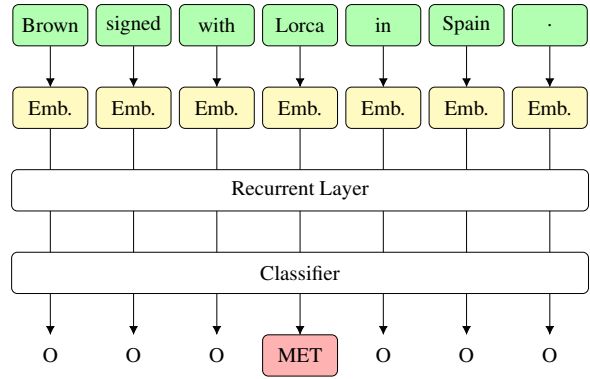
\begin{figure}
\resizebox{\linewidth}{!}{
\begin{tikzpicture}[node distance=2cm, align=center]
	\node (token1Node)		[tokenBlock]             							{Brown};
	\node (rep1Node)		[smallBlock, below of=token1Node]					{Emb.};
	\node (label1Node)		[negLabelBlock, below of=token1Node, yshift=-6cm]	{O};
	\draw[->]             	(token1Node) 	--  	(rep1Node);
	\draw[->]             	(rep1Node) 		--  	(label1Node);

	\node (token2Node)		[tokenBlock, right of=token1Node]             		{signed};
	\node (rep2Node)		[smallBlock, below of=token2Node]					{Emb.};
	\node (label2Node)		[negLabelBlock, below of=token2Node, yshift=-6cm]	{O};
	\draw[->]             	(token2Node) 	--  	(rep2Node);
	\draw[->]             	(rep2Node) 		--  	(label2Node);

	\node (token3Node)		[tokenBlock, right of=token2Node]             		{with};
	\node (rep3Node)		[smallBlock, below of=token3Node]					{Emb.};
	\node (label3Node)		[negLabelBlock, below of=token3Node, yshift=-6cm]	{O};
	\draw[->]             	(token3Node) 	--  	(rep3Node);
	\draw[->]             	(rep3Node) 		--  	(label3Node);

	\node (token4Node)		[tokenBlock, right of=token3Node]             		{Lorca};
	\node (rep4Node)		[smallBlock, below of=token4Node]					{Emb.};
	\node (label4Node)		[posLabelBlock, below of=token4Node, yshift=-6cm]	{MET};
	\draw[->]             	(token4Node) 	--  	(rep4Node);
	\draw[->]             	(rep4Node) 		--  	(label4Node);

	\node (token5Node)		[tokenBlock, right of=token4Node]             		{in};
	\node (rep5Node)		[smallBlock, below of=token5Node]					{Emb.};
	\node (label5Node)		[negLabelBlock, below of=token5Node, yshift=-6cm]	{O};
	\draw[->]             	(token5Node) 	--  	(rep5Node);
	\draw[->]             	(rep5Node) 		--  	(label5Node);

	\node (token6Node)		[tokenBlock, right of=token5Node]             		{Spain};
	\node (rep6Node)		[smallBlock, below of=token6Node]					{Emb.};
	\node (label6Node)		[negLabelBlock, below of=token6Node, yshift=-6cm]	{O};
	\draw[->]             	(token6Node) 	--  	(rep6Node);
	\draw[->]             	(rep6Node) 		--  	(label6Node);

	\node (token7Node)		[tokenBlock, right of=token6Node]             		{.};
	\node (rep7Node)		[smallBlock, below of=token7Node]					{Emb.};
	\node (label7Node)		[negLabelBlock, below of=token7Node, yshift=-6cm]	{O};
	\draw[->]             	(token7Node) 	--  	(rep7Node);
	\draw[->]             	(rep7Node) 		--  	(label7Node);

	\node (recNode)			[longBlock, below of=rep4Node]						{Recurrent Layer};
	\node (clsNode)			[longBlock, below of=recNode]						{Classifier};
\end{tikzpicture}
}
\caption{Architecture of the \seqlab model.
		 The input sentence is \emph{Brown signed with Lorca in Spain}, where \meto{Lorca} is metonymic.
		 This model processes all the tokens of the input simultaneously.
		 }
\label{fig:nomaskmodel}
\end{figure}

\begin{figure}
\resizebox{\linewidth}{!}{
\begin{tikzpicture}[node distance=2cm, align=center]
	\node (token1Node)		[tokenBlock]										{Brown};
	\node (rep1Node)		[smallBlock, below of=token1Node]					{Emb.};
	\draw[->]             	(token1Node) 	--  	(rep1Node);
	\node (stealth1Node)	[stealthBlock, below of=token1Node, yshift=-4cm]	{};
	\draw[->]             	(rep1Node) 	--  	(stealth1Node);

	\node (token2Node)		[tokenBlock, right of=token1Node]             		{signed};
	\node (rep2Node)		[smallBlock, below of=token2Node]					{Emb.};
	\draw[->]             	(token2Node) 	--  	(rep2Node);
	\node (stealth2Node)	[stealthBlock, below of=token2Node, yshift=-4cm]	{};
	\draw[->]             	(rep2Node) 	--  	(stealth2Node);

	\node (token3Node)		[tokenBlock, right of=token2Node]             		{with};
	\node (rep3Node)		[smallBlock, below of=token3Node]					{Emb.};
	\draw[->]             	(token3Node) 	--  	(rep3Node);
	\node (stealth3Node)	[stealthBlock, below of=token3Node, yshift=-4cm]	{};
	\draw[->]             	(rep3Node) 	--  	(stealth3Node);

	\node (token5Node)		[tokenBlock, right of=token4Node]  					{in};
	\node (rep5Node)		[smallBlock, below of=token5Node]					{Emb.};
	\draw[->]             	(token5Node) 	--  	(rep5Node);
	\node (stealth5Node)	[stealthBlock, below of=token5Node, yshift=-4cm]	{};
	\draw[->]             	(rep5Node) 	--  	(stealth5Node);

	\node (token6Node)		[tokenBlock, right of=token5Node]             		{Spain};
	\node (rep6Node)		[smallBlock, below of=token6Node]					{Emb.};
	\draw[->]             	(token6Node) 	--  	(rep6Node);
	\node (stealth6Node)	[stealthBlock, below of=token6Node, yshift=-4cm]	{};
	\draw[->]             	(rep6Node) 	--  	(stealth6Node);

	\node (token7Node)		[tokenBlock, right of=token6Node]             		{.};
	\node (rep7Node)		[smallBlock, below of=token7Node]					{Emb.};
	\draw[->]             	(token7Node) 	--  	(rep7Node);
	\node (stealth7Node)	[stealthBlock, below of=token7Node, yshift=-4cm]	{};
	\draw[->]             	(rep7Node) 	--  	(stealth7Node);

	\node (rec1Node)		[mid2Block, below of=rep2Node]						{Recurrent Layer};
	\node (rec2Node)		[mid2Block, below of=rep6Node]						{Recurrent Layer};

	\node (catNode)			[longBlock, below of=rep4Node, yshift=-2cm]			{Concatenation};
	\node (clsNode)			[longBlock, below of=catNode]						{Classifier};
	\draw[-]             	(catNode) 		--  	(clsNode);

	\node (label4Node)		[posLabelBlock, below of=clsNode, yshift=0.2cm]		{MET};
	\draw[->]             	(clsNode) 		--  	(label4Node);
\end{tikzpicture}
}
\caption{Architecture of the \seqlabbasen{n} model, where n denotes the number of context words.
		 The input sentence is \emph{Brown signed with Lorca in Spain}, where \meto{Lorca} is metonymic.
		 This model processes only one token of the input at a time.
		 The target word currently is \emph{Lorca}.
		 }
\label{fig:maskmodel}
\end{figure}
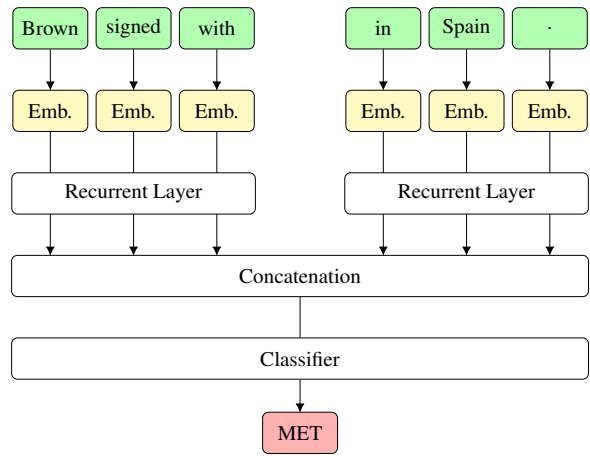

	\section{Experimental Setup}

	\subsection{Experimental Data}
	
	We conduct our experiments\footnote{The code will be made publicly available upon publication.}
		on \wiki\mycitep{lrec20-wimcor}\footnote{\url{https://github.com/nlpAThits/WiMCor}}.
	The other metonymy datasets,
		namely
		\semeval\mycitep{markert-nissim-2007-semeval}
		and 
		\relocar\mycitep{P17-1115},
		are small in size
		and
		hence inadequate
		for large-scale machine learning and statistical analyses.
	\wiki 
		is much larger in size
		and
		has richer annotation, 
		which leads to empirically more reliable and linguistically more meaningful results.
	The data
		is extracted
		automatically
		using 
		\wikipedia data.

	Every \wiki sample
		is composed of 
		a piece of text
		and
		two labels of varying granularity
		for the word designated as the \pmw.
 	The coarse-grained label 
		classifies the word
		into 
		\lit
		and
		\met.
	The fine-grained label
		classifies 
		the metonymic word
		further into
		entity types such as 
		\etype{Institution}, 
		\etype{Artifact}, 
		\etype{Team}
		and
		\etype{Event}.
	Note that 
		while metonymy operates
 		on different entity types
		such as locations, organizations or persons, 
		all the the positive instances 
		in \wiki
		are location names.
	For example, 
		the literal reading of 
		the token
		\meto{Barcelona}
		comprises 
		the geographical and locative interpretations 
		of the city of Barcelona in Spain.
	The token 
		is assigned 
		the label \lit 
		(see \exref{example:ex-lit}).
	If the same token is 
		used to denote 
		the football team 
		based in the city of Barcelona
		then the token is metonymic.
	The token
		is assigned
		the coarse-grained label \met,
		and
		the fine-grained label \etype{Team}
		(see \exref{example:ex-met}).		

	We modify the labels of \wiki
		to fit 
		the new task formulation.
	We consider
		unannotated words to be literal.
	\mycitet{gao-etal-2018-neural-metaphor}
		follow a similar approach
		in metaphor detection.
	However
		this assumption 
		introduces noise
		in the data
		because
		some words 
		that are not \pmw in the original version
		might be metonymic in reality.
	Such words
		are incorrectly labeled \lit
		in the modified version. 
%

	A statistical breakdown of the dataset
		after relabeling
		is shown in 
		\tabref{table:dataset-stats}
		and
		\tabref{table:dataset-tokens}.
	Note that the
		dataset is in English.
	The corpus
		is split
		into train, validation  and test partitions
		in the ratio 60:20:20 respectively.
	The metonymic mentions
		in each partition
		are pairwise disjoint
		with each other
		to ensure that the models
		treat metonymic
		and literal tokens alike.


%
				
	\begin{table}
		\centering
		{
		\begin{tabularx}{\linewidth}{  >{\raggedright\arraybackslash}l 
								|  >{\raggedleft\arraybackslash}X 
								|  >{\raggedleft\arraybackslash}X 
								}
		\hline
 		\textbf{Item}				
				& \textbf{Train}		
				& \textbf{Val/Test}		\\
	        \hline
		\# samples				
				& 123.6K
				& 41.2K				\\      
		\# sentences				
				& 478K
				& 159K				\\ 	
		\# avg. sample length		
				& 80
				& 80					\\
		\# tokens				
				& 12M
				& 4.1M				\\      
		\# unique tokens			
				& 338K
				& 189K				\\      
	        \hline
		\end{tabularx}
		\caption{
				Basic statistics of \wiki
				after relabeling the data 
				to fit sequence labeling. 
				}
		\label{table:dataset-stats}
		}
	\end{table}

	\begin{table}
		{
		\centering
		\begin{tabularx}{\linewidth}{  >{\raggedright\arraybackslash}l
								 | >{\raggedright\arraybackslash}l 
								 | >{\raggedleft\arraybackslash}X 
								 | >{\raggedleft\arraybackslash}X 
								 }
		\hline
 		\textbf{Coarse} 		
							& \textbf{Fine}	 	
							& \textbf{Train}		
							& \textbf{Val/Test}		
							\\
	        \hline
		{\small \lit} 					
							& 	\lit 				
							& 11.7M				
							& 4.08M
							\\	
		\hline
		\multirow{4}{*}{\small \met} 	
							& 	\etype{Institute}	
							& 18K				
							& 5.9K
							\\	

							& 	\etype{Artifact} 	
							& 6K					
							& 2.0K
							\\	

							& 	\etype{Team} 		
							& 6K					
							& 1.9K
							\\	

							& 	\etype{Event} 	
							& 1K					
							& 0.4K
							\\	
	        \hline
		\end{tabularx}
		\caption{
				The labels of tokens in \wiki
				after relabeling the data 
				to fit sequence labeling. 
				}
		\label{table:dataset-tokens}
		}
	\end{table}


	As we can observe from
		\figref{fig:nomaskmodel}
		and
		\figref{fig:maskmodel},
		while \seqlab model
		parses the whole input at once,
		\seqlabbasen{n}
		parses the input token by token.
	However
		the literal tokens 
		far outnumbers 
		the metonymic tokens
		with a a ratio of 1:377 (see \tabref{table:dataset-tokens}).
	So
		in the case of \seqlabbasen{n},
		we 
		randomly downsample the majority class	
		during the training phase
		to ensure that the classes are balanced.
	If the percentage of the majority class decreases, 
		then recall increases
		but precision decreases\mycitep{Zhang03}.
	The best balance between precision and recall
		is achieved 
		when about \negpercent\% of 
		the majority class is sampled.

	\subsection{Evaluation Settings}

	The objective
		of our experiments
		is to analyze how well
		 the models
		distinguish 
		metonymic usage of tokens 
		from 
		literal usage of tokens.
	We designed two different settings
		for evaluation:
		coarse-grained
		and
		fine-grained.
	The objective 
		of the former
		is to identify
		the metonymic tokens.
	The valid output label here
		is
		either
		\lit
		or
		\met.
	The objective
		of the latter
		is to identify
		the specific entity type
		referred to by a metonymic token.
	Hence 
		the valid output label
		is
		either 
		\lit for literal reading,
		or
		\etype{Institute},
		\etype{Artifact},
		\etype{Team}
		or
		\etype{Event}
		for metonymic reading.
%

 	\subsection{Evaluation Metrics}

	We evaluate each model
		using the following
		classification metrics:
		precision, 
		recall 
		and 
		F1-score.
	As the class distribution in the dataset is imbalanced (see \tabref{table:dataset-tokens}),
		we report micro-averaged and macro-averaged metrics.
	While macro-averaging 
		treats all classes alike, 
		micro-averaging 
		takes into account
		the proportion of each class in the data.
	Since we are interested in evaluating
		how well the models detect
		metonymic readings
		as opposed to 
		literal readings,
		we consider 
		the set of metonymic instances as the positive class.
	All the metrics reported in \secref{section:results}
		pertain to the positive class only.
	
	\subsection{Baselines}

	We use two baseline systems
		for comparison.

	\subsubsection{Probabilistic Modeling}
	\label{section:crf}
	

	In this paper, 
		we choose \crf 
		owing to its robust performance
		in various
		sequence labeling tasks
		such as
		entity analysis\mycitep{durrett-klein-2014-joint},
		part-of-speech tagging\mycitep{silfverberg-etal-2014-part},
		parsing\mycitep{finkel-etal-2008-efficient},
		and
		language modeling\mycitep{roark-etal-2004-discriminative}.
	\crfs
		jointly predict
		the labels of the entire sequence
		by taking into account 
		the transition probabilities of labels.	
	
	We use
		four sets of hand-crafted features to train this model:		
	\begin{asparaenum}[(1)]
		\item surface-level features --- whether the token has only alphabetic characters, 
			 whether the token has only digits, 
			 whether the token is punctuation,
			 whether the token has only uppercase characters,
			 and
			 whether the token starts with an uppercase character,
		\item syntactic features --- part-of-speech (POS) tag of a token,
		\item n-gram features --- token in lemmatized form, and
		\item grammatical features --- dependency role of a token,
			and
			the 2-tuple (dependency role, dependency head).
	\end{asparaenum}
	We compile
		this feature set
		from 
		previous work
		on metonymy resolution
		and
		commonly used features
		in other sequence labeling tasks.
	\mycitet{Markert:2002:MRC:1118693.1118720}
		conclude that
		generalized collocation features,
		as opposed to co-occurrence features,
		are useful for metonymy resolution.
	More recent work
		such as\mycitet{Nastase:2009:CCL:1699571.1699631} and\mycitet{Nastase:2012:LGC:2390948.2390971}
		focus on selectional preference features.
	Some of the features
		proposed
		in the metonymy literature
		do not fit
		our experimental settings.
	For instance,
		we do not use
		the features
		\emph{number of grammatical roles}
		and
		\emph{number of words}
		proposed by\mycitet{Nissim:2003:SFW:1075096.1075104}
		because
		the former characterizes only mixed readings 
		and
		the latter is rendered irrelevant
		as we assume every word to be potentially metonymic.

	\subsubsection{Context-insensitive Embeddings}

	We also use
		context-insensitive embeddings for comparison.
	For this  purpose, 
		we use
		50-dimensional
		\glove\mycitep{pennington-etal-2014-glove}
		trained on six billions tokens
		in Wikipedia 2014 and English Gigaword (fifth edition) corpus. 

\section{Results and Discussion}\label{section:results}

		
	The results
		of each model
		are reported in
	\tabref{table:models-results}.
	We experiment 
		with different context windows.
	All the models 
		exhibit 
		better performance
		in coarse-grained setting
		as compared to 
		fine-grained setting
		due to fewer number of classes
		and
		thus lesser dependence on the context for disambiguation.

		

	The feature set of \seqcrf
		consists of
		surface-level, syntactic and grammatical features of the current token,
		and 
		the surface-level, syntactic, grammatical features and lemmatized forms of the context words.
%
	We classify
		the top 1000 informative features 
		on the basis of the group to which they belong.
	We observe
		that 
		grammatical features (92.5\%) are the most informative,
		followed by 
		n-gram (5.6\%), surface (0.8\%) and syntactic (0.9\%) features respectively.

%


	\begin{table*}
		\centering
		{
		\begin{tabular}{  m{0.15\linewidth} 
					     | M{0.05\linewidth} M{0.05\linewidth} M{0.05\linewidth}
					     | M{0.05\linewidth} M{0.05\linewidth} M{0.05\linewidth}
					     | M{0.05\linewidth} M{0.05\linewidth} M{0.05\linewidth} 
					     }
		\hline
		
		\multirow{2}{*}{\textbf{Model}} 	 
			& \multicolumn{3}{ c | }{\textbf{Coarse-grained}}				
			& \multicolumn{6}{ c  }{\textbf{Fine-grained}}				\\
		\cline{5-10}
			& \multicolumn{3}{ c | }{\textbf{}}			
			& \multicolumn{3}{ c | }{\textbf{Micro-average}}			
			& \multicolumn{3}{ c  }{\textbf{Macro-average}}					\\
		\cline{2-10}
			& \textbf{P}			& \textbf{R}		& \textbf{F1}					
			& \textbf{P}			& \textbf{R}		& \textbf{F1}					
			& \textbf{P}			& \textbf{R}		& \textbf{F1}			\\
		\hline

			\seqcrf  
			& .59				& .27			& .37
			& .58				& .26			& .36
			& .39				& .17			& .23			
			\\
		\hline
			\seqlab{\textsubscript{\glove}}
			& .12			& .03		& .04
			& .11			& .01		& .02
			& .07			& .01		& .02
			\\
			\seqlabbasen{5}{\textsubscript{, \glove}}
			& .38			& .45		& .41
			& .42			& .30		& .35
			& .26			& .14		& .14
			\\
			\seqlabbasen{10}{\textsubscript{, \glove}}
			& .36			& .53		& .43
			& .40			& .41		& .41
			& .25			& .26		& .25
			\\
			\seqlabbasen{50}{\textsubscript{, \glove}}
			& .32			& .51		& .39
			& .36			& .48		& .41
			& .22			& .33		& .25
			\\

		\hline
			\seqlab{\textsubscript{\bert}}
			& .83			& .62		& \textbf{.71}
			& .81			& .56		& .66
			& .66			& .36		& .43				
			\\
			\seqlabbasen{5}{\textsubscript{, \bert}}
			& .58			& .85		& .69
			& .60			& .79		& \textbf{.69}
			& .57			& .68		& \textbf{.61}				
			\\
			\seqlabbasen{10}{\textsubscript{, \bert}}
			& .58			& .83		& .69
			& .62			& .77		& .69
			& .59			& .64		& .60
			\\
			\seqlabbasen{50}{\textsubscript{, \bert}}
			& .54			& .84		& .66
			& .66			& .68		& .67
			& .57			& .48		& .49
			\\

%
			

%
		\hline
		\end{tabular}
		}
		\caption{Performance of different models on \wiki data.
				The \seqlabbasen{n} models
				outperform
				the \seqlab model
				by a significant margin
				as shown by the macro-averaged
				metrics of the 
				fine-grained evaluation setting.
				}
		\label{table:models-results}
	\end{table*}

	As mentioned in \secref{section:introduction},
		while previous work
		treats
		metonymy detection 
		as a token-level classification task,
		in this paper, 
		we reformulate 
		metonymy detection
		as a sequence labeling task.
	In coarse-grained evaluation,
		\mycitet{li2020target}
		reports an accuracy of 95.9.
	On the other hand,
		even the best performing model
		in \tabref{table:models-results}
		achieves an F1-score of 0.71 only.

	\subsection{Impact of Target Word}

	In the context of metonymy detection,
		the target word
		can be used in different ways.
	\mycitet{Nastase:2012:LGC:2390948.2390971}
		and\mycitet{Nastase:2009:CCL:1699571.1699631}
		compute selectional preferences 
		of the target word
		for the words in its context.
	The model proposed by\mycitet{P17-1115}
		does not take into consideration
		the surface form of the target word.
	\mycitet{li2020target}
		mask the target word
		during training and evaluation.

	As shown in \tabref{table:models-results},
		the performances
		of \seqlab and \seqlabbasen{5}
		are similar to each other (0.71 and 0.69 F1-scores respectively)
		in the coarse-grained setting.
	This is because 
		coarse-grained setting
		considers
		all fine-grained entity types
		as \met
		and thus
		takes into account
		the proportion of each entity type in the dataset.
	For the same reason,
		in the
		fine-grained setting,
		the micro-averaged metrics 
		of \seqlab and \seqlabbasen{5}
		exhibit a similar behavior (0.66 and 0.69 F1-scores respectively).
	On the other hand,
		the macro-averaged metrics,
		which disregards the proportion of labels in the dataset,
		in the fine-grained evaluation setting
		clearly demonstrate
		the difference between
		the two models.
	For instance,
		while the macro-averaged F1-score
		of the \seqlab model
		is 0.43,
		the \seqlabbasen{5} model
		exhibits
		a much higher F1-score of 0.61.

	The results discussed above
		indicate that
		the target word
		is less useful 
		for end-to-end metonymy detection,
		since the masked variants
		outperform
		the unmasked variant.
	In addition,
		\seqlab{\textsubscript{\glove}},
		which relies on the target word and nothing else,
		is the worst performing of all the models.
	The irrelevance 
		of the target word
		can be attributed to 
		two factors.
	First,
		all the metonymic mentions
		in the dataset
		are location names, 
		which, in turn, are proper nouns. 
	Even if we replace
		the metonymic mention
		\emph{Barcelona} in \exref{example:ex-met}
		with any other location,
		the sentence will retain
		its figurative nature.
	So 
		the surface form
		of the mention
		is less relevant here.
	On the other hand,
		the target word
		might be more useful
		for detecting metonymy in 
		other word categories such as common nouns,
	For instance,
		consider the sentence
		``The ham sandwich is waiting for his cheque.''\mycitep{lakoff1980metaphors}.
	Here
		the noun phrase \emph{the ham sandwich} is metonymic
		because it refers to the customer who placed the order.
	If this phrase is replaced with the word \emph{customer},
		then the sentence is no longer figurative.

	Second, 
		in this paper,
		we
		study the impact of target word
		in metonymy detection,
		which
		only predicts whether the target word is metonymic or literal.
	The next phase in metonymy resolution
		is metonymy interpretation\mycitep{Teraoka2016MetonymyAU,Markert:2002:MRC:1118693.1118720},
		which identifies
		the \emph{hidden} entity 
		represented by the target word.
	We believe
		the surface form of the target word
		to be 
		more useful in 
		the interpretation phase
		as compared to 
		the detection phase.

	\subsection{Impact of Context Words}

	\begin{table*}[h!]
		\centering
		\parbox{\linewidth}
		{
		\centering
		\begin{tabular}
			{   >{\raggedright\arraybackslash}c
			  \myvrule{.1pt}  >{\raggedleft\arraybackslash}c     >{\raggedleft\arraybackslash}c     >{\raggedleft\arraybackslash}c 
			  \myvrule{.1pt}  >{\raggedleft\arraybackslash}c     >{\raggedleft\arraybackslash}c     >{\raggedleft\arraybackslash}c
			  \myvrule{.1pt}  >{\raggedleft\arraybackslash}c     >{\raggedleft\arraybackslash}c     >{\raggedleft\arraybackslash}c
			  \myvrule{.1pt}  >{\raggedleft\arraybackslash}c     >{\raggedleft\arraybackslash}c     >{\raggedleft\arraybackslash}c
			  }
		\hline
 		\multirow{2}{*}{\textbf{Class}}				
 			& \multicolumn{3}{c \myvrule{.1pt} }{\textbf{\seqlab}}
 			& \multicolumn{3}{c|}{\textbf{\seqlabbasen{5}}}
 			& \multicolumn{3}{c|}{\textbf{\seqlabbasen{10}}}
 			& \multicolumn{3}{c}{\textbf{\seqlabbasen{50}}}
 			\\
	        \cline{2-13}
 			& {P}	& {R}	& {F1}
 			& {P}	& {R}	& {F1}
 			& {P}	& {R}	& {F1}
 			& {P}	& {R}	& {F1}
 			\\
                 \hline
		\etype{Team}		
			& .89		& {.49}		& .63
			& .65		& .89		& .75
			& .69 		& .88 		& \textbf{.77}
			& .77 & .74 & .75
			\\      
		\etype{Institute}
			& .84		& .77		& \textbf{.80}
			& .69		& .91		& .78
			& .71 		& .89 		& .79
			& .72 & .84 & .78
			\\      
		\etype{Artifact}			
			& .33		& {.08}		& .13
			& .31		& .41		& \textbf{.36}
			& .29 		& .37 		& .32
			& .30 & .23 & .26
			\\      
		\etype{Event}				
			& .58		& {.10}		& .17
			& .64		& .51		& \textbf{.57}
			& .70 		& .40 		& .51
			& .51 &.12 & .19
			\\      
	        \hline
		\end{tabular}
		\caption{Per-class performance
				of different models.
				Only the models based on \bert
				are shown here for comparison.
				All the models exhibit {better} performance 
				on \etype{Team} and \etype{Institute},
				compared to
				\etype{Artifact} and \etype{Event}.
				}
		\label{table:per-class-results}
		}
	\end{table*}

				
	\begin{table*}[h!]
		\centering
		{
		\begin{tabular}{ l | l | l | l | l }
		\hline
		\textbf{Type}
			& \textbf{\etype{Team}}
			& \textbf{\etype{Institute}}
			& \textbf{\etype{Artifact}}
			& \textbf{\etype{Event}}
			\\
		\hline
		\multirow{10}{*}{\shortstack{\textbf{Top 10}\\\textbf{context}\\\textbf{words}}}
		         &	club		     	&	university		    &	raf		     		&	battle
		     	\\
		         &	fc		     		&	college		     	&	squadron		    &	tokugawa
		     	\\
		         &	match		     	&	professor		    &	cathedral			&	clan
		     	\\
		         &	league		     	&	degree		     	&	airport		     	&	ieyasu
		     	\\
		         &	goal		     	&	teach		     	&	prix				&	domain
		     	\\
		         &	season		     	&	science     		&	fly		     		&	province
		     	\\
		         &	cup		     		&	institute		    &	aircraft		    &	norman
		     	\\
		         &	loan		     	&	school		     	&	length		     	&	festival
		     	\\
		         &	score		     	&	study		     	&	bury		     	&	fight
		     	\\
		         &	uefa		     	&	honorary			&	stakes		     	&	regiment
		     	\\
		\hline
		    \shortstack{\textbf{Avg. cosine}\\\textbf{similarity}}
			&	0.62	&	\textbf{0.69}
			&	0.33  	&	0.34
	     	\\

%
		\hline
		\end{tabular}
		}
		\caption{
			Top 10 words
				(taken from context window of 5 tokens each from either side)
				associated with the mentions of each class
				in descending order of normalized PMI scores.
			Average cosine similarity	
				of each word list
				(computed using
				\glove 50-dimensional word embeddings)
				is shown at the bottom
				of the table.
			\etype{Team}
			and
			\etype{Institute},
			are associated with many 
			domain-specific words 
			(and hence higher average cosine similarity),
			compared to
			\etype{Artifact}
			and
			\etype{Event}.
			}
		\label{table:pmi}
	\end{table*}

	The per-class performance
		of the 
		best variant of each model
		is shown in \tabref{table:per-class-results}.
	The F1-scores
		of
		the entity types \etype{Team} and \etype{Institute}
		are much higher 
		compared to 
		that of
		\etype{Artifact} and \etype{Event}.
	We hypothesize 
		that context words
		have high impact
		on the classification,
		and
		might throw light on
		the disparity
		in performance
		on different entity types.
	It has previously been noted that
		context words
		in the form of 
		collocation and 
		cooccurrences
		are relevant to 
		metonymy detection\mycitep{Markert:2002:MRC:1118693.1118720}.

	To better understand the impact of context, 
		we analyze the data
		using the context words
		of mentions 
		belonging to each class.
	For this purpose,
		we measure
		the association between context words and classes
		using normalized pointwise mutual information (\pmi)\mycitep{church-hanks-1989-word}.
	Normalized \pmi
		handles the bias of \pmi
		towards low frequency data\mycitep{Bouma2009}.
	First 
		we parse the train partition of our data
		to extract the immediate context words 
		of all positive 
		mentions.
	We 
		convert the words into their lemmatized form 
		and
		filter out 
		words
		whose count falls below a threshold of 2.
%
	Normalized \pmi is then computed as follows:
	$$ 
		\text{PMI}(w, c) = \text{log} \frac{\text{p}(w, c)}{\text{p}(w)\text{p}(c)};  
	$$
	$$
		\text{NPMI}(w, c) = \frac{\text{PMI}(w, c)}{-\text{log}(w, c)};  
	$$
	where 
	$w$ and $c$
	correspond 
	to context word and class respectively.
	
	\tabref{table:pmi} 
		shows
		the top 10 words
		(taken from context window of 5 tokens each from either side)
		associated with
		each class
		in descending order of normalized \pmi scores.
	While
		the context words 
		associated with
		\etype{Team} and \etype{Institute}
		are domain-specific in nature, 
		the context words 
		associated with 
		\etype{Artifact} and \etype{Event}
		are more general in nature.

	We also report the average cosine similarity
		between the context words
		of each entity type.
	We use
		the 50-dimensional 
		\glove
		embeddings
		to 
		compute the similarity between 
		each pair of context words.
	The similarity scores
		are then aggregated 
		through simple averaging
		to compute 
		the average cosine similarity
		of each entity type.
	As shown in \tabref{table:pmi} ,
		the average cosine similarities of 
		\etype{Institute} and \etype{Team}
		are significantly higher 
		compared to that of 
		\etype{Event} and \etype{Artifact}.
	For instance,
		the top context words
		associated with \etype{Institute}
		such as
		\emph{university}, 
		\emph{college},
		\emph{professor}
		and
		\emph{degree}
		are highly representative of educational institutions
		and 
		thus have a high average cosine similarity of 0.69.
	On the other hand,
		the top context words
		associated with \etype{Artifact}
		such as
		\emph{raf} (which stands for Royal Air Force stations), 
		\emph{squadron},
		\emph{cathedral}
		and
		\emph{airport}
		exhibit more diversity
		and 
		thus have a low average cosine similarity of 0.33.


	These results 
		demonstrate 
		that
		the models
		fail to perform well
		when the mention
		is flanked by 
		domain-agnostic context words.
	This observation
		reaffirms
		the significance of context
		for detecting metonymy in text.

	\subsection{Impact of Window Size}

	Following\mycitet{gritta-etal-2018-melbourne},
		we experiment
		with contexts 
		of different sizes: 5, 10 and 50.
	In case of the \glove embeddings,
		short context maximises precision but lowers recall
		because
		the model misses out on important words lying outside the context window.
	Long context maximises recall but lowers precision
		because
		the model encounters irrelevant words.
	For instance, 
		as shown in \tabref{table:models-results},
		the macro-averaged precision
		drops from 0.26 to 0.22
		as the window size is increased from 5 to 50, 
		and
		the macro-averaged recall
		rises from 0.14 to 0.33
		as the window size is increased from 5 to 50.
	
	On the other hand,
		since the \bert embeddings
		are context-sensitive, 
		increasing the context window
		has a negative impact on performance.
	However, 
		as shown in \tabref{table:per-class-results},
		\seqlabbasen{5}
		outperforms 
		both
		\seqlabbasen{10}
		and
		\seqlabbasen{50}
		for the minority class \etype{Event}
		suggesting that 
		shorter window size
		facilitates faster learning.


\subsection{Error Analysis}

	\begin{table*}[ht!]
		{
		\centering
		\begin{tabularx}{\linewidth}{ l | m{7.5cm} | l | l | l | l | l }
		\hline
		\textbf{No.}
			& \textbf{Example}
			& \textbf{Label}
			& \rotatebox{70}{\textbf{\seqlab}}
			& \rotatebox{70}{\textbf{\seqlabbasen{5}}}
			& \rotatebox{70}{\textbf{\seqlabbasen{10}}}
			& \rotatebox{70}{\textbf{\seqlabbasen{50}}}
		\\
		\hline
		(1) &
		In January 2018, Brown signed with \meto{Lorca} in Spain.
		& \etype{Team}
		& \checkmark  & \checkmark  & \checkmark & \checkmark \\

		\hline
		(2) &
		He was Born in Vigo, Galicia, Spain on 16 February 1958. In 1982 Virgós graduated from \meto{Compostela} with a degree in psychology. His training and professional activity were mainly in the  field of community intervention and social services.
		& \etype{Institute}
		& $\times$ & \checkmark  & \checkmark   & \checkmark \\

		\hline
		(3) &
		Blanco believed it better to fight than surrender to the Americans. He ordered Pascual Cervera y Topete to break the American blockade, leading to \textbf{Santiago}.
		& \etype{Event}
		& $\times$ & \checkmark & $\times$ & $\times$ \\

		\hline
		(4) &
		Lawrence has also worked on \meto{Reading}'s educational website Romans Revealed, creating stories about Roman Britain closely based on archaeological finds.
		& \etype{Institute}
		& $\times$ & $\times$ & $\times$ & $\times$ \\

		\hline
			\end{tabularx}
		\caption{
			Error analysis.
			The metonymic mention is 
			shown in boldface.
			Only the models based on \bert
			are shown here for comparison.
			}
		\label{table:erran}
		}
	\end{table*}

	To further understand the 
		difference between 
		\seqlab and \seqlabbasen{n} models,
		we conduct 
		a manual error analysis. 
	We compiled 
		four examples (see \tabref{table:erran})
		representing
		the manner in which
		the models
		treat the samples in the dataset.
	In this subsection,
		we discuss
		each example
		in detail.

	\begin{asparaenum}[(1)]
	\item
	Both 
		\seqlab  and \seqlabbasen{n} models
		solve this example correctly.
	The predicate \emph{`sign with'}
		is a strong indicator of the metonymic sense
		of the target word \emph{Lorca}.

	\item
	\seqlab  fails to solve this example 
		because the metonymic mention \emph{Compostela} is metonymic in 
		\etype{Artifact}, 
		\etype{Institute}, 
		and 
		\etype{Location} train partitions. 
	The model prediction 
		is highly tied to the current string 
		thus and 
		however the \seqlabbasen{n} models 
		rely on context for disambiguation.
	For instance, 
		the name Heidelberg can refer to,
		among others, 
		the Heidelberg Castle (\etype{Artifact}), 
		the Heidelberg University (\etype{Institute}) 
		or 
		the city Heidelberg (\etype{Location}). 
	A metonymy detection system
		should be able to distinguish 
		these interpretations from the context.

	\item
	While
		\seqlabbasen{10}
		misclassifies the metonymic mention \emph{Santiago}
		as \etype{Artifact}
		and
		\seqlabbasen{50}
		fails to detect the mention
		as metonymic,
		\seqlabbasen{5}
		correctly identifies the mention
		as an \etype{Event}.

	\item
	All the models
		fails to even detect the metonymic mention.
	This example 
		is difficult to resolve 
		due to a lack of any domain-specific word 
		in the neighbourhood of the metonymic mention.
	\end{asparaenum}

\section{Conclusions}
Previous work
	treats
	metonymy detection 
	as a token-level classification task,
	which 
	disambiguates
	only a single pre-specified mention in the input.
In this paper, 
	we reformulate 
	metonymy detection
	as a sequence labeling task
	by disambiguating
	every word in the input.
We show that
	the new setting
	is
	computationally harder
	compared to 
	token-level classification.

We
	also
	investigate the impact of target word and context
	on metonymy detection.
The model
	that relies primarily on context
	outperforms 
	the model that has access to both context and the target word
	in the fine-grained classification task.
This shows that
	the target word
	is less useful
	for detecting metonymy
	in \wimcor.
On the other hand,
	the entity types
	that are associated with
	domain-specific words
	in their context
	are easier to solve.
This shows that
	the context words are
	much more relevant
	for detecting metonymy.


In order to fully resolve metonymy,	
	it is necessary to
	identify
	the specific \emph{hidden} entity
	referred to by the metonymic mention.
This interpretation task
	can be formulated as entity linking,
	which attempts to
	link
    named entity mentions in text
    to entities in a knowledge base.
\wimcor,
	unlike 
	the other metonymy datasets 
	\relocar and \semeval, 
	is suitable 
	for this task
	because
	the metonymic mentions in \wimcor
	are linked to \wikipedia.


\bibliography{master}
\bibliographystyle{acl_natbib}

\end{document}


\appendix

\section{Downsampling Majority Class}\label{appendix:downsample}

While \seqlab model
	parses the whole input at once,
	\seqlabbasen{n}
	parses the input token by token.
However,
	the literal tokens 
	far outnumbers 
	the metonymic tokens.
So, 
	in the case of \seqlabbasen{n},
	we 
	randomly downsample the majority class	
	during the training phase
	to ensure that the classes are balanced.

The classification metrics 
	with respect to
	the size of the majority class
	in terms of percentage of sampled units
	are shown in
	\figref{figure:seqlabbase}.
If the percentage of the majority class decreases, 
	then recall increases
	but precision decreases.
Without resampling, 
	the recall is very low.
The best balance between precision and recall
	is achieved 
	when about \negpercent\% of 
	the majority class is sampled.

\begin{figure}[ht!]
  	\centering
	\resizebox{0.5\textwidth}{!}{
		\input{images/seqlabbase-coarse.pdf_tex}
   	 }
	\caption{
	Comparison of 
	evaluation metrics
	of the \seqlabbase model
	with respect to 
	percentage of negative examples for resampling.
	The optimal F1-score
	is achieved when \negpercent\%  of the negative class is sampled.}
    \label{figure:seqlabbase}
\end{figure}

\section{\semeval}

	\begin{table}
		\centering
		{
		\begin{tabularx}{\linewidth}{  >{\raggedright\arraybackslash}l
								|  >{\raggedleft\arraybackslash}X
								|  >{\raggedleft\arraybackslash}X
								}
		\hline
 		\textbf{Item}				& \textbf{Train}	& \textbf{Val/Test}		\\
        \hline
		\# samples					& 1078				& 359					\\
		\# sentences				& 1083				& 360					\\
		\# avg. sample length		& 34				& 34					\\
		\# tokens					& 36K				& 12K					\\
		\# unique tokens			& 8.6K				& 3.8K					\\
        \hline
		\end{tabularx}
		\caption{Basic statistics of \semeval
			after relabeling the data
			to fit sequence labeling.
		}
		\label{table:semeval-stats}
		}
	\end{table}

  \begin{table}
    \centering
    {
    \begin{tabular}{  m{0.30\linewidth} 
               | M{0.15\linewidth} M{0.15\linewidth} M{0.15\linewidth}
               }
    \hline

    \multirow{2}{*}{\textbf{Model}}
      & \multicolumn{3}{ c }{\textbf{Coarse-grained}}  \\
    \cline{2-4}
      & \textbf{P}      & \textbf{R}    & \textbf{F1}  \\
    \hline
      \seqlab{\textsubscript{\glove}}
      & .00     & .00   & .00      \\
      \seqlabbasen{5}{\textsubscript{, \glove}}
      & .00     & .00   & .00      \\
      \seqlabbasen{10}{\textsubscript{, \glove}}
      & .00     & .00   & .00      \\
      \seqlabbasen{50}{\textsubscript{, \glove}}
      & .00     & .00   & .00      \\
    \hline
      \seqlab{\textsubscript{\bert}}
      & .00     & .00   & .00      \\
      \seqlabbasen{5}{\textsubscript{, \bert}}
      & .16     & .10   & .12      \\
      \seqlabbasen{10}{\textsubscript{, \bert}}
      & .28     & .09   & .14      \\
      \seqlabbasen{50}{\textsubscript{, \bert}}
      & .25     & .21   & .23      \\
    \hline
    \end{tabular}
    }
	    \caption{Performance of different models on \semeval data.
        }
    \label{table:semeval-results}
  \end{table}

\pagebreak
\section{\relocar}

	\begin{table}
		\centering
		{
		\begin{tabularx}{\linewidth}{  >{\raggedright\arraybackslash}l
								|  >{\raggedleft\arraybackslash}X
								|  >{\raggedleft\arraybackslash}X
								}
		\hline
 		\textbf{Item}				& \textbf{Train}		& \textbf{Val/Test}		\\
        \hline
		\# samples					& 1200					& 400					\\
		\# sentences				& 1200					& 400					\\
		\# avg. sample length		& 22					& 22					\\
		\# tokens					& 30K					& 10K					\\
		\# unique tokens			& 7.6K					& 3.2K					\\
        \hline
		\end{tabularx}
		\caption{Basic statistics of \relocar
			after relabeling the data
			to fit sequence labeling.
		}
		\label{table:relocar-stats}
		}
	\end{table}

  \begin{table}
    \centering
    {
    \begin{tabular}{  m{0.30\linewidth} 
               | M{0.15\linewidth} M{0.15\linewidth} M{0.15\linewidth}
               }
    \hline

    \multirow{2}{*}{\textbf{Model}}    
      & \multicolumn{3}{ c }{\textbf{Coarse-grained}}  \\
    \cline{2-4}
      & \textbf{P}      & \textbf{R}    & \textbf{F1}  \\
    \hline
      \seqlab{\textsubscript{\glove}}
      & .08     & .01   & .02      \\
      \seqlabbasen{5}{\textsubscript{, \glove}}
      & .17     & .14   & .14      \\
      \seqlabbasen{10}{\textsubscript{, \glove}}
      & .15     & .23   & .18      \\
      \seqlabbasen{50}{\textsubscript{, \glove}}
      & .17     & .19   & .17      \\
    \hline
      \seqlab{\textsubscript{\bert}}
      & .25     & .07   & .11      \\
      \seqlabbasen{5}{\textsubscript{, \bert}}
      & .40     & .32   & .35      \\
      \seqlabbasen{10}{\textsubscript{, \bert}}
      & .35     & .38   & .37      \\
      \seqlabbasen{50}{\textsubscript{, \bert}}
      & .35     & .44   & .39      \\
    \hline
    \end{tabular}
    }
    \caption{Performance of different models on \relocar data.
        }
    \label{table:relocar-results}
  \end{table}